\documentclass[sigconf,screen,nonacm]{acmart}  
\AtBeginDocument{%
  }

\setcopyright{acmlicensed}
\copyrightyear{2026}
\acmYear{2026}
\acmDOI{XXXXXXX.XXXXXXX}
\acmConference[MM '26]{34th ACM International Conference on Multimedia}{November 10--14, 2026}{Rio de Janeiro, Brazil}
\acmISBN{978-1-4503-XXXX-X/2026/11}

\acmSubmissionID{1096}

\usepackage[table]{xcolor}  
\usepackage[utf8]{inputenc}
\input{laeenc.def}
\makeatletter
\newcount\c@chapter \c@chapter=\z@
\makeatother
\usepackage[arabic,english]{babel}
\makeatletter
\renewcommand{\thesection}{\@arabic\c@section}
\renewcommand{\thesubsection}{\thesection.\@arabic\c@subsection}
\renewcommand{\thesubsubsection}{\thesubsection.\@arabic\c@subsubsection}
\renewcommand{\thefigure}{\@arabic\c@figure}
\renewcommand{\thetable}{\@arabic\c@table}
\renewcommand{\theequation}{\@arabic\c@equation}
\renewcommand{\thefootnote}{\@arabic\c@footnote}
\makeatother
\usepackage{tikz}
\usepackage{url}
\usepackage{enumitem}  
\usepackage{multirow,makecell}  
\usepackage{arydshln}
\usepackage{colortbl}  
\usepackage{pifont,bbding}  
\usepackage{xspace}
\usepackage{array}  
\definecolor{SciGreen}{RGB}{44, 160, 44} 
\definecolor{SciRed}{RGB}{214, 39, 40}
\newcommand{\cmark}{\textcolor{SciGreen}{\ding{51}}}
\newcommand{\xmark}{\textcolor{SciRed}{\ding{55}}}

\newcolumntype{G}{>{\columncolor{gray!10}}c}
\newcommand{\gc}{\cellcolor{gray!10}}

\newcommand{\circledi}{
  \raisebox{-0.12ex}{\tikz[baseline=(char.base)]{
    \node[shape=circle,draw,inner sep=0.8pt] (char) {\hspace{-1pt}\textit{i}};
  }}
}
\newcommand{\circledzero}{
  \tikz[baseline=(char.base)]{
    \node[shape=circle,draw,inner sep=0.7pt] (char) {\small{0}};
  }
}

\begin{document}

\title{Habibi: Laying the Open-Source Foundation of Unified-Dialectal Arabic Speech Synthesis}

\author{Yushen Chen}
\orcid{0009-0002-5477-7352}
\affiliation{
  \institution{Shanghai Jiao Tong University}
  \institution{Shanghai Innovation Institute}
  \city{Shanghai}
  \country{China}
}
\email{swivid@sjtu.edu.cn}

\author{Junzhe Liu}
\orcid{0009-0000-1293-8723}
\affiliation{
  \institution{Shanghai Jiao Tong University}
  \city{Shanghai}
  \country{China}
}

\author{Yujie Tu}
\orcid{0009-0008-3156-4833}
\affiliation[obeypunctuation=true]{
  \institution{University of the Chinese Academy of Sciences,}
  \city{Beijing,}
  \country{China\\}
  \institution{Shanghai Innovation Institute}
  \city{Shanghai,}
  \country{China}
}

\author{Zhikang Niu}
\orcid{0009-0007-1880-7434}
\affiliation{
  \institution{Shanghai Jiao Tong University}
  \institution{Shanghai Innovation Institute}
  \city{Shanghai}
  \country{China}
}

\author{Yuzhe Liang}
\orcid{0009-0006-2596-2814}
\affiliation{
  \institution{Shanghai Jiao Tong University}
  \institution{Shanghai Innovation Institute}
  \city{Shanghai}
  \country{China}
}

\author{Chunyu Qiang}
\orcid{0009-0007-2290-3074}
\affiliation{
  \institution{Kuaishou Technology}
  \city{Beijing}
  \country{China}
}

\author{Chen Zhang}
\orcid{0009-0006-5301-8695}
\affiliation{
  \institution{Kuaishou Technology}
  \city{Beijing}
  \country{China}
}

\author{Kai Yu}
\orcid{0000-0002-7102-9826}
\affiliation{
  \institution{Shanghai Jiao Tong University}
  \city{Shanghai}
  \country{China}
}

\author{Xie Chen}
\authornote{Corresponding author.}
\orcid{0000-0001-7423-617X}
\affiliation{
  \institution{Shanghai Jiao Tong University}
  \institution{Shanghai Innovation Institute}
  \city{Shanghai}
  \country{China}
}

\renewcommand{\shortauthors}{Yushen Chen et al.}

\begin{abstract}
Arabic spans over 30 spoken varieties, yet no open-source text-to-speech system unifies them. Key barriers include substantial cross-dialect lexical and phonological divergence, scarce synthesis-grade data, and the absence of a standardized multi-dialect evaluation benchmark. We present \textbf{Habibi}, a unified-dialectal Arabic TTS framework that addresses all three. Through a multi-step curation pipeline, we repurpose open-source ASR corpora into TTS training data covering 12+ regional dialects. A linguistically-informed curriculum learning strategy—progressing from Modern Standard Arabic to dialectal data—enables robust zero-shot synthesis without text diacritization. We further release the first standardized multi-dialect Arabic TTS benchmark, comprising over 11,000 utterances across 7 dialect subsets with manually verified transcripts. On this benchmark, our unified model matches or surpasses per-dialect specialized models. Both automatic metrics and human evaluations confirm that Habibi is highly competitive with ElevenLabs' Eleven v3 (alpha) in intelligibility, speaker similarity, and naturalness. Extensive ablations (${\sim}$8,000 H100 GPU hours, 30+ configurations) validate each design choice. We open-source all checkpoints, training and inference code, and benchmark data—the first such release for multi-dialect Arabic TTS—at \url{https://SWivid.github.io/Habibi/}.
\end{abstract}



\begin{teaserfigure}
  \centering
  \includegraphics[width=0.98\textwidth]{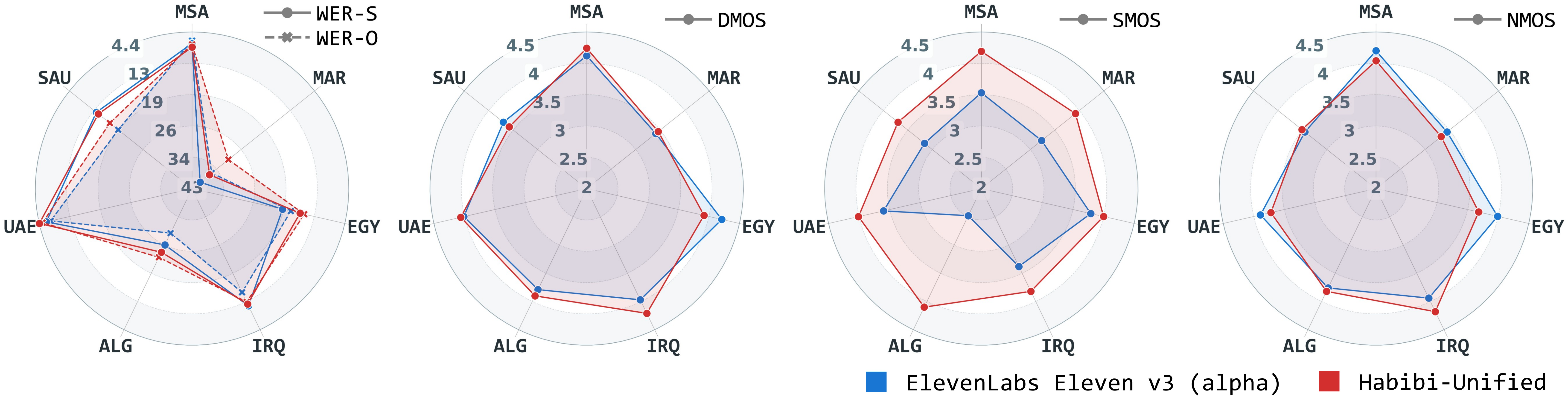}
  \caption{Our open-source unified-dialectal model (Habibi-Unified) vs.\ ElevenLabs' commercial Eleven v3 (alpha). WER-S/O: word error rates from two different ASR systems. D/S/N-MOS: dialect pronunciation accuracy, speaker similarity, and naturalness.}
  \Description{The radar charts compare zero-shot TTS performance between Habibi and ElevenLabs' commercial service across seven dialects in the benchmark, evaluated along four dimensions. Habibi matches ElevenLabs in intelligibility and naturalness while significantly outperforming it in speaker similarity.}
  \label{fig:compared_with_11labs}
\end{teaserfigure}


\maketitle


\section{Introduction}
Arabic is a pluricentric language with rich regional diversity~\cite{pluricentric,dialectal}, spoken natively by over 400 million people in approximately 30 distinct varieties~\cite{enwiki:1329477423}. From a speech technology perspective, the inherent challenges of Arabic lie in two aspects: (1) \textit{Linguistically}, the Modern Standard Arabic (MSA), as the official and literary form, is rarely used in everyday life, while the various spoken dialects used in daily communication undergo extensive and ongoing fusion and shift. Different dialects exhibit distinct lexical and phonological conventions, resulting in considerable vocabulary and grammatical complexity.
(2) \textit{Technically}, the challenge is that although MSA is high-resource, many dialects remain underrepresented and are considered low-resource. Across dialects, text diacritization can help indicate correct pronunciation. However, existing labeled data and in-the-wild transcriptions alike often lack diacritics. Moreover, most existing datasets are designed for automatic speech recognition (ASR) and contain a large proportion of noisy samples. Virtually no clean, ready-made datasets exist at scale for speech synthesis.

Current trends in Arabic dialectal speech research span two main directions: shifting ASR focus beyond MSA toward low-resource dialects (MGB-2, MGB-3, MGB-5 challenges~\cite{mgb2,mgb3,mgb5}, \textit{etc.}), and efforts to expand dialectal coverage of Arabic ASR~\cite{dialectal,nadi2025,omnilingual}; and advancing text-to-speech (TTS) from MSA-centric systems~\cite{asc,clartts,artst} to a handful of models with limited dialect support~\cite{tunartts,arzstts}, whose results still fall short of those achieved by recent zero-shot TTS models for languages such as English and Chinese.
The underwhelming performance of these models is unsurprising given the limitations of existing Arabic speech datasets. For instance, ArVoice~\cite{arvoice}, by far the largest open-source corpus specifically designed for TTS, contains fewer than 100 hours of MSA-only data from 11 speakers, much of which is synthetic. Meanwhile, the QASR corpus~\cite{qasr}, primarily intended for ASR, features a highly imbalanced gender distribution among its speakers. This imbalance, though likely unintentional, renders the corpus unsuitable for training high-quality TTS models.
Beyond data scarcity, there remains little impetus for developing Arabic dialectal TTS, a common issue for long-tail languages, driven by scientific disincentives and other factors~\cite{omnilingual}.
Notably, to the best of our knowledge, research on unified-dialectal Arabic TTS remains absent, let alone an open-source framework.

To address all aforementioned challenges, we present our \textbf{Habibi} ({\small \AR{حبيبي}}, “my dear friend” here):
\begin{itemize}[leftmargin=*, topsep=0pt]
    \item The first open-source unified-dialectal TTS, covering more than 20 languoids (ISO 639-3 codes) and 12 assignable regional identifiers to date, competitive with the leading commercial model.
    \item The first standardized benchmark for multi-dialect Arabic zero-shot speech synthesis, featuring 1,000 to 3,000 utterances in each of 7 subsets, paired with manually verified and filtered transcripts.
    \item A linguistically-informed curriculum learning strategy, progressing from high- to low-resource data and from general to dialect-aware training, enables the model to capture subtle dialectal features via in-context learning at inference without requiring diacritized text, advancing beyond a proof of concept.
    \item An extensively ablated and scalable solution for multi- and unified-dialectal Arabic TTS, through which we deliver insights into the pros and cons of training a single unified model versus separate specialized models per dialect. We further demonstrate the importance of incorporating dialect-specific ASR models for more comprehensive and reliable evaluation.
\end{itemize}

\section{Methodology}
\label{subsec:methodology}

This section presents the Habibi framework. We first define the abbreviation conventions for Arabic variants (\S\ref{subsec:terminology}), then describe our multi-dialect data curation pipeline (\S\ref{subsec:dataset}) and benchmark (\S\ref{subsec:benchmark}). 
Finally, we detail the two key training strategies: linguistically-informed curriculum learning (\S\ref{subsec:staged_training}) and dialect-aware supervised fine-tuning (\S\ref{subsec:region_id}).

\subsection{Terminology}
\label{subsec:terminology}

In the absence of a strict scientific definition of Arabic dialects, we adopt a practical convention, referring to Arabic variants at two levels of granularity. We emphasize that this convention \textit{does not reflect any official classification}. The two levels are: languoids (linguistic entities from ISO 639-3) and regional identifiers (IDs) based on countries or regions (three capitalized letters).

\textbf{\texttt{MSA}}\quad  Modern Standard Arabic. This is an official variant of Arabic, commonly used in literary form, \textit{e.g.}, in news, books, and education. Coded \texttt{arb\_Arab} in ISO 639-3.

\textbf{\texttt{SAU}}\quad  Referring to the various spoken dialects of Arabic in Saudi Arabia, primarily Najdi (Central, \texttt{ars\_Arab}) and Hijazi (Western, \texttt{acw\_Arab}), Khaliji or Gulf Arabic (Eastern, \texttt{afb\_Arab}), with more including Baharna (\texttt{abv\_Arab}).

\textbf{\texttt{UAE}}\quad  Emirati Arabic, a group of Gulf Arabic (\texttt{afb\_Arab}) varieties spoken by the Emiratis native to the United Arab Emirates.

\textbf{\texttt{ALG}}\quad  Dialects spoken in Algeria, including primarily Algerian Arabic (\texttt{arq\_Arab}), and Algerian Saharan Arabic (\texttt{aao\_Arab}).

\textbf{\texttt{IRQ}}\quad  Iraqi Arabic, mainly spoken in the Mesopotamian basin of Iraq as well as parts of Syria (\texttt{acm\_Arab} \& \texttt{ayp\_Arab}).

\textbf{\texttt{EGY}}\quad  Egyptian Arabic (\texttt{arz\_Arab}) is the most widely spoken vernacular Arabic variety in Egypt. Saidi Arabic (\texttt{aec\_Arab}), or Upper Egyptian Arabic, is another spoken variety in the south.

\textbf{\texttt{MAR}}\quad  Moroccan Arabic (\texttt{ary\_Arab}), or Darija, is spoken by the majority of people in Morocco.

\textbf{\texttt{OMN}}\quad  Including Omani Arabic (\texttt{acx\_Arab}) and Dhofari Arabic (\texttt{adf\_Arab}), in our data sources.

\textbf{\texttt{TUN}}\quad  Tunisian Arabic (\texttt{aeb\_Arab}), or simply Tunisian, is a variety of Arabic spoken in Tunisia.

\textbf{\texttt{LEV}}\quad  Levantine Arabic (\texttt{apc\_Arab}), spoken in the Levant, also includes Levantine Bedawi Arabic (\texttt{avl\_Arab}) here.

\textbf{\texttt{SDN}}\quad  Sudanese Arabic (\texttt{apd\_Arab}) refers to the various varieties of Arabic spoken in Sudan.

\textbf{\texttt{LBY}}\quad  Libyan Arabic (\texttt{ayl\_Arab}) is a variety of Arabic spoken in Libya and neighboring countries.

\subsection{Multi-Dialect Data Curation Pipeline}
\label{subsec:dataset}

\begin{table}[ht]
\centering
\caption{Curated dataset statistics. $\dagger$ indicates expanded data for training; \texttt{UNK} marks unknown or non-applicable dialect.}
\label{tab:dataset_source}
\setlength{\tabcolsep}{8pt}
\resizebox{\columnwidth}{!}{%
\begin{tabular}{llrr}
\toprule
\textbf{Dataset Name} & \textbf{ID}~~~~~ & \textbf{Hours} & \textbf{Utterances} \\
\midrule
MASC\texttt{[MSA]}~\cite{masc} & \texttt{MSA} & 329.0 & 152,405 \\
\midrule
MGB-2~\cite{mgb2} & \texttt{MSA} & 585.2 & 215,958 \\
\midrule
SADA~\cite{sada}  & \texttt{SAU} & 187.0 & 154,288 \\
                  & \texttt{MSA} & 7.2 & 4,090 \\
                  & \texttt{EGY} & 2.1 & 2,029 \\
                  & \texttt{LEV} & 0.9 & 947 \\
                  & \texttt{UNK} & 219.2 & 79,807 \\
\midrule
Mixat~\cite{mixat}             & \texttt{UAE} & 13.1 & 4,746 \\
\midrule
UAE-100K       & \texttt{UAE} & 94.4 & 50,035 \\
\midrule
UAE-Nexdata    & \texttt{UAE} & 0.6 & 423 \\
\midrule
MASC\texttt{[EGY]}~\cite{masc} & \texttt{EGY} & 18.5 & 15,910 \\
\midrule
MGB-3~\cite{mgb3}              & \texttt{EGY} & 8.3 & 3,165 \\
\midrule
FLEURS~\cite{fleurs}           & \texttt{EGY} & 8.2 & 2,824 \\
\midrule
MGB-5~\cite{mgb5}              & \texttt{MAR} & 33.7 & 26,382 \\
\midrule
in-house & \texttt{ALG} & 64.4 & 70,970 \\
         & \texttt{IRQ} & 58.9 & 48,534 \\
         & \texttt{UAE} & 4.4 & 2,413 \\
\midrule
Omnilingual ASR Corpus${}^\dagger$~\cite{omnilingual} & \texttt{SAU} & 62.4 & 8,819 \\
        & \texttt{ALG} & 8.6 & 1,482 \\
        & \texttt{IRQ} & 11.7 & 1,726 \\
        & \texttt{EGY} & 25.1 & 3,979 \\
        & \texttt{MAR} & 8.3 & 1,659 \\
        & \texttt{OMN} & 20.7 & 2,778 \\
        & \texttt{TUN} & 19.4 & 2,632 \\
        & \texttt{LEV} & 2.5 & 383 \\
        & \texttt{SDN} & 4.2 & 612 \\
        & \texttt{LBY} & 14.8 & 2,194 \\
\midrule
Darija-S2T${}^{\dagger}$ & \texttt{MAR} & 30.6 & 3,839 \\
\midrule
DarijaTTS-clean${}^{\dagger}$ & \texttt{MAR} & 9.2 & 10,292 \\
\midrule
Jordan-Audio${}^{\dagger}$ & \texttt{LEV} & 4.9 & 4,016 \\
\midrule
\textbf{Total} & & \textbf{1857.3} & \textbf{879,339} \\
\bottomrule
\end{tabular}%
}
\end{table}

\begin{table}[ht]
\centering
\caption{Duration statistics comparing basic data (D1) and data after expansion (D2) across different dialects.}
\label{tab:datasets_dialect}
\setlength{\tabcolsep}{8pt}
\resizebox{0.9\columnwidth}{!}{
\begin{tabular}{lrr}
\toprule
\textbf{Regional} & \textbf{Basic Data (hrs)} & \textbf{After Expansion (hrs)} \\
\textbf{Identifier} & \textbf{D1} as \textit{abbr.} & \textbf{D2} as \textit{abbr.} \\
\midrule
\texttt{MSA} & 921.5 & 921.5 \\
\texttt{SAU} & 187.0 & 249.3 \\
\texttt{UAE} & 112.4 & 112.4 \\
\texttt{ALG} & 64.4  & 72.9 \\
\texttt{IRQ} & 58.9  & 70.7 \\
\texttt{EGY} & 37.1  & 62.2 \\
\texttt{MAR} & 33.7  & 81.7 \\
\texttt{OMN} & -     & 20.7 \\
\texttt{TUN} & -     & 19.4 \\
\texttt{LEV} & 0.9   & 8.3 \\
\texttt{SDN} & -     & 4.2 \\
\texttt{LBY} & -     & 14.8 \\
\texttt{UNK} & 219.2 & 219.2 \\
\midrule
\textbf{Total} & \textbf{1635.0} & \textbf{1857.3} \\
\bottomrule
\end{tabular}
}
\end{table}

We assembled training data from several open-source ASR datasets: MASC~\cite{masc} (MSA and Egyptian parts), SADA~\cite{sada}, MGB-2~\cite{mgb2}, MGB-3~\cite{mgb3}, MGB-5~\cite{mgb5}, FLEURS~\cite{fleurs}, and Omnilingual ASR Corpus~\cite{omnilingual}.
We also collected publicly available datasets: UAE-100K\footnote{\url{https://huggingface.co/datasets/AhmedBadawy11/UAE_100K}}, UAE-Nexdata\footnote{\url{https://huggingface.co/datasets/Nexdata/UAE_Arabic_Spontaneous_Speech_Data}}, Darija-S2T\footnote{\url{https://huggingface.co/datasets/adiren7/darija_speech_to_text}}, DarijaTTS-clean\footnote{\url{https://huggingface.co/datasets/KandirResearch/DarijaTTS-clean}}, and Jordan-Audio\footnote{\url{https://huggingface.co/datasets/nadsoft/Jordan-Audio}}, where the source links are provided in Table~\ref{tab:dataset_source}.
We finally supplemented these sources with internal data and manually transcribed public speech recordings.

Since the raw data were entirely uncurated and likely contaminated with significant noise detrimental to TTS training, we implemented a multi-step selective processing procedure as follows.

For all datasets, we use speaking rate (Characters Per Second, CPS) as a filtering criterion. 
This is based on several observations: an excessively small CPS often indicates missing text labels or audio with long silent segments, while an overly large CPS suggests incorrect automatic subtitles or corrupted short audio files remaining in the original datasets. 
Though simple, this filter is demonstrably effective: manual verification confirms it removes a substantial portion of noisy instances. The thresholds are empirically determined, with lower bounds in $[4, 10]$ and upper bounds in $[15, 25]$, adjusted per dataset. The guiding principle is consistent: the higher the estimated corruption, the more aggressive the thresholds.

For low-SNR datasets, we employed a source separation model~\cite{bsrnn} to denoise speech, discarding samples rendered silent post-separation. The efficacy of this approach entails a nuanced trade-off (\S\ref{subsec:ablation_data}).

A meticulous filtering procedure was applied exclusively to the MASC corpus~\cite{masc}, for two reasons. First, MASC constitutes a significant portion of the total data. Second, because many MGB-2 clips were recorded at effective sampling rates below 8 kHz, it was crucial to ensure that the higher-sampling-rate MASC was properly filtered to reliably support the model's learning of MSA, the major Arabic variant.
Concretely, we used only the \textit{clean} subset. Through manual inspection, we established text-pattern-based filtering rules—since MASC was crawled from YouTube channels, corruption patterns were highly channel-specific (\textit{e.g.}, recurring intro/outro music or singing). After rule-based filtering, we merged adjacent segments shorter than 6 seconds into clips of up to 30 seconds, shifting the duration distribution from the concentrated sub-10-second range toward lengths better suited for TTS.

Final dataset statistics are in Table~\ref{tab:dataset_source} and \ref{tab:datasets_dialect}, where D1 and D2 denote Basic Data and Expanded Data, respectively.

\subsection{Multi-Dialect Arabic TTS Benchmark}
\label{subsec:benchmark}

To our knowledge, no standardized benchmark exists for multi-dialect Arabic TTS evaluation. 
We establish the first benchmark for the multi-dialect Arabic zero-shot TTS task, covering 7 Arabic variants: \texttt{MSA}, \texttt{SAU}, \texttt{UAE}, \texttt{ALG}, \texttt{IRQ}, \texttt{EGY}, and \texttt{MAR}. All test sets undergo the same data preparation pipeline described in \S\ref{subsec:dataset}. We then select samples meeting three criteria: (1) duration between 3 and 12 seconds; (2) transcriptions in Arabic script only; and (3) at least one other utterance from the same speaker exists—a requirement for zero-shot TTS evaluation, which needs both reference and target speech per speaker.

For each released subset, the source and size are as follows:

\textbf{\texttt{MSA}}\quad 2902 samples from MGB-2 \textit{test} set;

\textbf{\texttt{SAU}}\quad 1375 samples from SADA \textit{val} and \textit{test} sets;

\textbf{\texttt{UAE}}\quad 1048 samples from UAE-100K \textit{val} set;

\textbf{\texttt{ALG}}\quad 1727 samples from in-house data;

\textbf{\texttt{IRQ}}\quad 2198 samples from in-house data;

\textbf{\texttt{EGY}}\quad 1024 samples from MGB-3 \textit{test} set;

\textbf{\texttt{MAR}}\quad 1083 samples from MGB-5 \textit{test} set.

We ensure strict separation between benchmark and training data, and release the full benchmark for reproducible evaluation.

\subsection{Linguistically-Informed Curriculum Learning}
\label{subsec:staged_training}

We employ a two-stage curriculum learning strategy to enhance generative performance across Arabic dialects. In the first stage, we initialize from F5-TTS~\cite{f5tts}, pre-trained on approximately 95K hours of Chinese and English data (\S\ref{subsec:training_setup}). We then perform supervised fine-tuning (SFT) on MSA data only. 
The rationale is linguistic: as the formal written standard shared across all Arabic-speaking regions, MSA provides the most regular phonological and grammatical patterns, making it an ideal bridge for transferring the model's text-to-speech capability from Chinese and English to Arabic. This stage allows the model to acquire foundational Arabic articulation before encountering the greater variability of dialectal data.

The first-stage endpoint is determined empirically by selecting the checkpoint where the naturalness metric (\textit{e.g.}, UTMOS) converges to its optimum. 
This is because, while pronunciation accuracy (\textit{e.g.}, WER) continues improving with further training, the generated audio shifts toward the narrower SFT data distribution—largely from ASR corpora with lower SNR—degrading speaker similarity and naturalness. We therefore prioritize naturalness convergence as the transition criterion.

In the second stage, we pursue two parallel strategies (ablated in \S\ref{subsec:effect_stage_training}): specialized models fine-tuned on individual dialects, and unified models fine-tuned on all dialect data jointly. The former targets peak per-dialect performance, while the latter tests whether a single model can generalize across the full dialectal spectrum—a central question of this work.


\subsection{Dialect-Aware Supervised Fine-Tuning}
\label{subsec:region_id}

Current mainstream speech synthesis models exhibit strong in-context learning capabilities. When both reference audio and paired transcription are provided as conditions, these models can implicitly retrieve and effectively leverage text-to-acoustic correspondence from known pairs during zero-shot inference (\S\ref{subsec:effect_in_context_learning}). However, an optional regional identifier could further help clarify generation behavior and assist disambiguation.

We therefore augment the dictionary with special tokens for regional identifiers\circledi (the $i$-th term in \S\ref{subsec:terminology}), yielding text sequence:
\begin{equation}
\circledi, ~~\langle~, ~c_1, ~c_2, ~\ldots, ~c_j, ~\ldots, ~c_n, ~~\rangle~,
\end{equation}
where $c_j$ is the $j$-th character of the raw text of total length $n$, $\langle\cdot\rangle$ wraps a certain dialectal text with two special tokens indicating the beginning and ending.

As shown in \S\ref{subsec:ablation_region_id}, explicitly incorporating regional identifiers during training improves inference performance—even when identifiers are absent at test time. We hypothesize two factors. First, the curriculum-trained base model already possesses robust in-context learning capabilities and resilience to diverse input patterns. Second, training with identifiers helps the model internalize dialect-related distributional patterns more effectively.

\section{Experiments}
\label{sec:experiments}
In this section, we elaborate the experimental setup (\S\ref{subsec:training_setup}--\ref{subsec:eval_setup}) and present our main results (\S\ref{subsec:11labs_compr_results}--\ref{subsec:result_habibi}). 
Beyond the primary comparisons, we invest approximately 8,000 NVIDIA H100 SXM GPU hours in a systematic suite of ablation studies, each targeting a specific research question that underpins the design of Habibi: (1) Is linguistically-informed curriculum learning necessary, and how much does the MSA-only training stage contribute compared to direct dialectal fine-tuning? (\S\ref{subsec:effect_stage_training}) (2) To what extent does the model rely on in-context learning from reference speech-text pairs to capture dialectal features at inference time? (\S\ref{subsec:effect_in_context_learning}) (3) How do utterance length diversity, data quality, and overall dataset scale individually affect synthesis performance? (\S\ref{subsec:ablation_data}) (4) Do explicit regional identifiers yield measurable gains, and how robust is the model to different inference templates? (\S\ref{subsec:ablation_region_id}) Together, these experiments—spanning over 30 model configurations across 7 dialects—validate each core component of our framework and provide actionable insights for future work on low-resource dialectal speech synthesis.

\subsection{Backbone Choice and Training Setup}
\label{subsec:training_setup}

Current paradigms for zero-shot TTS primarily fall into three categories: 
autoregressive (AR) models~\cite{tacotron,valle,melle,sparktts,llasa,fishspeech,sac,calm}, 
non-autoregressive (NAR) models~\cite{fastspeech,fastspeech2,e3tts,ns3,voicebox,e2tts,maskgct,f5tts,zipvoice,m3tts}, 
and hybrid AR-NAR systems~\cite{tortoisetts,xtts,cosyvoice,seedtts,fireredtts,indextts2,ditar,minimaxspeech,distar,joyvoice}.
To our knowledge, there is no clear evidence that audio and text encoder modules prevalent in AR or hybrid systems can effectively capture the complex features of Arabic dialects without additional training. Given the need to avoid cascading errors, we adopted the open-source F5-TTS framework~\cite{f5tts}, which operates directly on mel spectrograms and raw text character sequences. This approach is not only straightforward but has also been validated in mainstream languages such as Chinese and English; its potential for extension to a wider range of languages is reflected in recent works~\cite {taskvector,lemas} and within the broader speech community.

Regarding configurations, we strictly follow the default F5-TTS v1 base training setup. To ensure adequate convergence across dialects, we train all models to 200K updates, with each run taking approximately 2 days on 8 NVIDIA H100 SXM GPUs.

\subsection{Evaluation Setup}
\label{subsec:eval_setup}

\begin{table*}[ht]
\centering
\caption{Reference audio sample entries with corresponding transcriptions across different Arabic regional identifiers, used for comparison with ElevenLabs. $\dagger$ indicates an ElevenLabs' PVC voice, while the others are from the Habibi benchmark.}
\label{tab:11labs_compr_reference_prompts}
\resizebox{\textwidth}{!}{
\renewcommand{\arraystretch}{1.5}
\begin{tabular}{llr}
\toprule
\textbf{ID} & \textbf{ElevenLabs Voice ID / Habibi Benchmark Entry} & \textbf{Transcription} \\
\midrule
\textbf{\texttt{MSA}} & \texttt{JjTirzdD7T3GMLkwdd3a}${}^\dagger$ & \AR{كان اللعيب حاضرًا في العديد من الأنشطة والفعاليات المرتبطة بكأس العالم، مما سمح للجماهير بالتفاعل معه والتقاط الصور التذكارية.} \\
\midrule
\multirow{3}{*}{\textbf{\texttt{SAU}}} 
& \texttt{6k\_SBA\_22\_2\-seg\_410\_975\-416\_518 (Najdi)} & \AR{تكفى طمني انا اليوم ماني بنايم ولا هو بداخل عيني النوم الين اتطمن عليه.} \\
\cdashline{2-3}
& \texttt{6k\_SBA\_27\_0\-seg\_544\_950\-550\_440 (Hijazi)} & \AR{ابغاك تحقق معاه بس بشكل ودي لانه سلطان يمر بظروف صعبة شوية.} \\
\cdashline{2-3}
& \texttt{6k\_SBA\_107\_1\-seg\_22\_960\-29\_010 ~(Gulf)} & \AR{وين تو الناس متى تصحى ومتى تفطر وتغير يبيلك ساعة يعني بالله تروح الشغل الساعة عشره.} \\
\midrule
\textbf{\texttt{UAE}} & \texttt{13\_segment\_108} & \AR{قمنا نشتريها بشكل متكرر أو لما نلقى ستايل يعجبنا وحياناً هذا الستايل ما نحبه.} \\
\midrule
\textbf{\texttt{ALG}} & \texttt{yCfWAGx6aTw\-162} & \AR{أنيا هكا باغية ناكل هكا أني ن نشوف فيها الحاجة هذيكا.} \\
\midrule
\textbf{\texttt{IRQ}} & \texttt{P9zYhlu5pzw\-613} & \AR{يعني ااا ما نقدر ناخذ وقت أكثر، ااا لأنه شروط كلش يحتاجلها وقت.} \\
\midrule
\textbf{\texttt{EGY}} & \texttt{IES4nrmZdUBHByLBde0P}${}^\dagger$ & \AR{ايه الكلام. بقولك ايه. استخدم صوتي في المحادثات. استخدمه هيعجبك اوي.} \\
\midrule
\textbf{\texttt{MAR}} & \texttt{OfGMGmhShO8iL9jCkXy8}${}^\dagger$ & \AR{إذا بغيتي شي صوت باللهجة المغربية للإعلانات ديالك هذا أحسن واحد غادي تلقاه.} \\
\bottomrule
\end{tabular}
}
\end{table*}

All experiments adhere to the benchmarking standards defined in \S\ref{subsec:benchmark}. We measure three conventional metrics: word error rate (WER) using ASR models, speaker similarity (SIM) leveraging the speaker verification model WavLM~\cite{wavlm}, and naturalness with UTMOS~\cite{utmos}. 
Notably, we report two sets of WER scores: (1) \textbf{WER-O}, evaluated with the Omnilingual-ASR-LLM-7B model~\cite{omnilingual} (v1 with a fixed batch size of 64); and (2) \textbf{WER-S}, derived from dialect-specific ASR models, most of which trained following VietASR~\cite{vietasr} except for \texttt{EGY} and \texttt{MAR} (for which two XLSR fine-tuned models on Hugging Face\footnote{\url{https://huggingface.co/IbrahimAmin/egyptian-arabic-wav2vec2-xlsr-53}}$^,$\footnote{\url{https://huggingface.co/boumehdi/wav2vec2-large-xlsr-moroccan-darija}} are employed).
The rationale for introducing both is to enable more reliable conclusions: a multilingual model risks cross-dialect recognition bias (\textit{e.g.}, incorrectly ``correcting'' speech that lacks certain dialect features), whereas specialized models often suffer from poor generalization and noise robustness.

To ensure a fair and unbiased comparison with the leading commercial system, ElevenLabs' Eleven v3 (alpha), which imposes a strict monthly quota on custom voice uploads, making evaluation across all speaker timbres in the Habibi benchmark prohibitively expensive, we adopted the following protocol to select a single representative reference audio for each dialect subset. 
First, we identified candidate voices from ElevenLabs' official PVC voice library\footnote{\url{https://elevenlabs.io/docs/creative-platform/voices/voice-cloning}}, selecting the most frequently used voice for each target dialect.
For each candidate, we performed secondary validation using a large language model to confirm that the audio preview exhibited authentic dialect-specific characteristics based on its transcription (extracted using ElevenLabs' Scribe v2). Once validated, the audio was designated as the reference for comparative evaluation. If no suitable voice was available in ElevenLabs' library, we applied the same protocol to audio samples from our Habibi benchmark. All selected ElevenLabs Voice IDs and Habibi benchmark entries (uploaded as IVC\footnote{\url{https://elevenlabs.io/docs/api-reference/voices/ivc/create}} when calling the ElevenLabs API), along with texts, are provided in Table~\ref{tab:11labs_compr_reference_prompts}. 
Note that the \texttt{SAU} subset employs distinct entries for the Najdi, Hijazi, and Gulf Arabic varieties.

We further conducted rigorous controlled subjective evaluations to verify the practical usability of our models, motivated by the inherent limitations of the objective metrics. Specifically, the ASR models employed for WER computation remain weak in Arabic dialect settings (particularly for \texttt{ALG} and \texttt{MAR}, where even SOTA ASR models yield high WER for ground truth), while SIM and UTMOS may suffer from out-of-domain generalization issues. To address this, we designed subjective evaluations along three dimensions using Mean Opinion Score (MOS) protocols: \textbf{DMOS} for dialect pronunciation accuracy, \textbf{SMOS} for speaker similarity, and \textbf{NMOS} for overall naturalness. For each dialect--dimension pair (21 in total), 10 to 20 paid native speakers rated randomly sampled outputs from all benchmarked TTS models over 10 rounds.

\subsection{Comparison Results with ElevenLabs}
\label{subsec:11labs_compr_results}
Full subjective and objective metrics comparing ElevenLabs' TTS service and our models are presented in Table~\ref{tab:11labs_compr_subj_results} and \ref{tab:11labs_compr_obj_results}.

\begin{table}[ht]
\centering
\caption{Subjective evaluation results of ElevenLabs' Eleven v3 (alpha) (11Labs-3a) and Habibi specialized and unified models (Special. and Uni.D2-I).}
\label{tab:11labs_compr_subj_results}
\setlength{\tabcolsep}{6pt}
\resizebox{\linewidth}{!}{
\begin{tabular}{lccccccc}
\toprule
\textbf{Model} & \textbf{\texttt{MSA}} & \textbf{\texttt{SAU}} & \textbf{\texttt{UAE}} & \textbf{\texttt{ALG}} & \textbf{\texttt{IRQ}} & \textbf{\texttt{EGY}} & \textbf{\texttt{MAR}} \\
\midrule
\multicolumn{8}{c}{\textbf{DMOS} $\uparrow$} \\
\midrule
11Labs-3a & \underline{4.12} & \underline{3.70} & \underline{4.01} & \underline{3.79} & 3.97 & \textbf{4.21} & 3.41 \\
\hdashline
Special.  & 3.76 & \textbf{3.77} & 3.92 & 3.68 & \underline{4.06} & 3.62 & \textbf{3.52} \\
\rowcolor{gray!10}
Uni.D2-I  & \textbf{4.24} & 3.58 & \textbf{4.06} & \textbf{3.90} & \textbf{4.21} & \underline{3.92} & \underline{3.46} \\
\midrule
\multicolumn{8}{c}{\textbf{SMOS} $\uparrow$} \\
\midrule
11Labs-3a & 3.53 & 3.16 & 3.60 & 2.48 & 3.38 & \underline{3.79} & 3.23  \\
\hdashline
Special.  & \underline{3.56} & \underline{3.69} & \underline{3.74} & \underline{3.78} & \textbf{3.82} & 3.12 & \underline{3.58}  \\
\rowcolor{gray!10}
Uni.D2-I  & \textbf{4.19} & \textbf{3.70} & \textbf{4.01} & \textbf{4.10} & \textbf{3.82} & \textbf{4.00} & \textbf{3.92}  \\
\midrule
\multicolumn{8}{c}{\textbf{NMOS} $\uparrow$} \\
\midrule
11Labs-3a & \textbf{4.20} & 3.45 & \textbf{3.89} & \underline{3.76} & \underline{3.94} & \textbf{3.99} & \textbf{3.45}  \\
\hdashline
Special.  & 3.89 & \textbf{3.56} & \underline{3.85} & 3.74 & 3.83 & 3.38 & \underline{3.44}  \\
\rowcolor{gray!10}
Uni.D2-I  & \underline{4.04} & \underline{3.51} & 3.72 & \textbf{3.82} & \textbf{4.18} & \underline{3.68} & 3.33  \\
\bottomrule
\end{tabular}
}
\end{table}

\begin{table}[ht]
\centering
\caption{Objective evaluation results of ElevenLabs' Eleven v3 (alpha) (11Labs-3a) and Habibi models. Special.: dialect-specialized models; Uni.D1/D2-I: unified models trained on D1/D2, with regional identifiers (``-I'').}
\label{tab:11labs_compr_obj_results}
\setlength{\tabcolsep}{3.5pt}
\resizebox{\linewidth}{!}{
\begin{tabular}{lccccccc}
\toprule
\textbf{Model} & \textbf{\texttt{MSA}} & \textbf{\texttt{SAU}} & \textbf{\texttt{UAE}} & \textbf{\texttt{ALG}} & \textbf{\texttt{IRQ}} & \textbf{\texttt{EGY}} & \textbf{\texttt{MAR}} \\
\midrule
\multicolumn{8}{c}{\textbf{WER-O} $\downarrow$} \\
\midrule
11Labs-3a & ~~\textbf{6.77} & 18.94 & ~~7.61 & 29.39 & 14.95 & 17.60 & 35.79 \\
\hdashline
Special.  & 10.37 & \underline{16.89} & ~~\textbf{5.16} & \underline{23.57} & \textbf{12.26} & 19.23 & 36.39 \\
Uni.D1-I  & ~~8.20 & 17.99 & ~~5.36 & 24.82 & 12.98 & \underline{15.19} & \textbf{30.05} \\
\rowcolor{gray!10}
Uni.D2-I  & ~~\underline{7.83} & \textbf{16.87} & ~~\underline{5.33} & \textbf{22.96} & \underline{12.85} & \textbf{14.96} & \underline{30.13} \\
\midrule
\multicolumn{8}{c}{\textbf{WER-S} $\downarrow$} \\
\midrule
11Labs-3a & ~~\textbf{7.54} & \underline{13.57} & ~~6.38 & 26.03 & \underline{11.69} & 19.27 & 40.04 \\
\hdashline
Special.  & 10.91 & \textbf{13.11} & ~~\underline{4.47} & \textbf{23.27} & \textbf{10.79} & 24.02 & 40.38 \\
Uni.D1-I  & ~~8.94 & 14.31 & ~~4.54 & 25.34 & 12.39 & \underline{17.40} & \underline{36.67} \\
\rowcolor{gray!10}
Uni.D2-I  & ~~\underline{8.56} & 14.07 & ~~\textbf{4.44} & \underline{24.21} & 12.23 & \textbf{15.79} & \textbf{36.58} \\
\midrule
\multicolumn{8}{c}{\textbf{SIM} $\uparrow$} \\
\midrule
11Labs-3a & 0.567 & 0.490 & 0.615 & 0.306 & 0.572 & 0.528 & 0.615 \\
\hdashline
Special.  & 0.698 & 0.657 & 0.854 & \textbf{0.744} & 0.811 & 0.511 & 0.629 \\
Uni.D1-I  & \textbf{0.811} & \underline{0.702} & \underline{0.859} & \underline{0.738} & \textbf{0.826} & \underline{0.654} & \underline{0.754} \\
\rowcolor{gray!10}
Uni.D2-I  & \underline{0.809} & \textbf{0.705} & \textbf{0.861} & 0.731 & \underline{0.825} & \textbf{0.686} & \textbf{0.757} \\
\midrule
\multicolumn{8}{c}{\textbf{UTMOS} $\uparrow$} \\
\midrule
11Labs-3a & \textbf{3.35} & \textbf{2.80} & \textbf{3.33} & \textbf{2.46} & \textbf{2.74} & \textbf{2.99} & \textbf{3.33} \\
\hdashline
Special.  & 1.84 & \underline{2.41} & \underline{2.92} & \underline{1.59} & \underline{2.59} & 2.68 & \underline{3.14} \\
Uni.D1-I  & 2.80 & 2.33 & 2.77 & 1.49 & 2.28 & 2.83 & 3.07 \\
\rowcolor{gray!10}
Uni.D2-I  & \underline{2.81} & 2.32 & 2.80 & 1.50 & 2.30 & \underline{2.92} & 3.08 \\
\bottomrule
\end{tabular}
}
\end{table}

\textbf{Overall competitiveness}\quad Our unified model (Uni.D2-I) match-es or surpasses 11Labs-3a on the majority of evaluated metrics. In subjective evaluation (Table~\ref{tab:11labs_compr_subj_results}), Uni.D2-I achieves higher DMOS on \texttt{MSA} (4.24 vs.\ 4.12), \texttt{UAE} (4.06 vs.\ 4.01), \texttt{ALG} (3.90 vs.\ 3.79), \texttt{IRQ} (4.21 vs.\ 3.97), and \texttt{MAR} (3.46 vs.\ 3.41), while outperforming 11Labs-3a on SMOS across all seven dialects, often by a substantial margin (\textit{e.g.}, \texttt{ALG}: 4.10 vs.\ 2.48). Objectively (Table~\ref{tab:11labs_compr_obj_results}), Uni.D2-I achieves lower WER-O on six of seven dialects and lower WER-S on four of seven, with SIM consistently higher across all dialects.

\textbf{Speaker similarity}\quad The most striking gap lies in speaker similarity. 11Labs-3a exhibits notably lower SIM scores (\textit{e.g.}, \texttt{ALG}: 0.306 vs.\ 0.731), corroborated by the subjective SMOS results where 11Labs-3a scores below 3.0 on \texttt{ALG} (2.48). This suggests that the ElevenLabs system may prioritize output naturalness over faithful reproduction of the reference speaker's voice characteristics.

\textbf{Naturalness vs.\ fidelity trade-off}\quad 11Labs-3a achieves higher UTMOS scores across all dialects (\textit{e.g.}, \texttt{MSA}: 3.35 vs.\ 2.81; \texttt{UAE}: 3.33 vs.\ 2.80), and leads in NMOS on most dialects. Combined with its lower SIM, this pattern is suspected to be a generation strategy that favors perceptual quality over strict speaker adherence. 
Note that the reference audio samples for \texttt{MSA}, \texttt{EGY}, and \texttt{MAR} are drawn from ElevenLabs' PVC voice library (Table~\ref{tab:11labs_compr_reference_prompts}), where 11Labs-3a was fine-tuned on over 30 minutes of same-speaker data; the resulting NMOS advantage over our zero-shot approach is thus expected.

\subsection{Benchmarking Habibi Models}
\label{subsec:result_habibi}

\begin{table}[ht]
\centering
\caption{Performance comparison of dialect-specific models across different training updates (Upd.).}
\label{tab:results_specific}
\setlength{\tabcolsep}{2.5pt}
\resizebox{\linewidth}{!}{
\begin{tabular}{lccccc}
\toprule
\textbf{ID~~~~~~~~} & ~~~\textbf{Upd.}~~~ & \textbf{WER-O} $\downarrow$ & \textbf{WER-S} $\downarrow$ & ~~\textbf{SIM} $\uparrow$~ & \textbf{UTMOS} $\uparrow$ \\ 
\midrule
\multirow{4}{*}{\texttt{MSA}} 
 & GT   & 11.42 & 10.92 & 0.802 & 2.11 \\
 \cdashline{2-6}
 & 100K & ~~8.43  & ~~7.85  & \textbf{0.764} & \textbf{2.01} \\
 & 150K & ~~7.95  & ~~\textbf{7.67}  & 0.763 & 1.98 \\
 & \gc200K & \gc~~\textbf{7.88}  & \gc~~7.74  & \gc0.763 & \gc1.93 \\
\midrule
\multirow{4}{*}{\texttt{SAU}} 
 & GT   & 28.36 & 19.88 & 0.724 & 1.71 \\
 \cdashline{2-6}
 & 100K & 14.47 & 11.12 & 0.711 & \textbf{1.79} \\
 & 150K & 13.62 & \textbf{10.36} & \textbf{0.712} & 1.75 \\
 & \gc200K & \gc\textbf{13.56} & \gc10.42 & \gc0.710 & \gc1.71 \\
\midrule
\multirow{4}{*}{\texttt{UAE}} 
 & GT   & 12.62 & 11.80 & 0.757 & 2.63 \\
 \cdashline{2-6}
 & \gc100K & \gc~~\textbf{4.97}  & \gc~~\textbf{4.55}  & \gc0.817 & \gc\textbf{2.88} \\
 & 150K & ~~5.40  & ~~5.03  & \textbf{0.821} & 2.77 \\
 & 200K & ~~5.73  & ~~5.49  & 0.820 & 2.66 \\
\midrule
\multirow{4}{*}{\texttt{ALG}} 
 & GT   & 41.19 & 26.57 & 0.754 & 2.27 \\
 \cdashline{2-6}
 & \gc100K & \gc\textbf{30.98} & \gc\textbf{19.85} & \gc0.740 & \gc\textbf{2.63} \\
 & 150K & 32.25 & 19.91 & \textbf{0.746} & 2.54 \\
 & 200K & 33.57 & 20.12 & \textbf{0.746} & 2.48 \\
\midrule
\multirow{4}{*}{\texttt{IRQ}} 
 & GT   & 27.18 & 18.05 & 0.790 & 2.42 \\
 \cdashline{2-6}
 & \gc100K & \gc\textbf{17.66} & \gc11.45 & \gc0.763 & \gc\textbf{2.63} \\
 & 150K & 18.50 & \textbf{11.42} & 0.769 & 2.54 \\
 & 200K & 19.33 & 11.60 & \textbf{0.770} & 2.47 \\
\midrule
\multirow{4}{*}{\texttt{EGY}} 
 & GT   & 22.70 & 18.62 & 0.820 & 2.20 \\
 \cdashline{2-6}
 & \gc100K & \gc\textbf{18.08} & \gc15.42 & \gc\textbf{0.769} & \gc\textbf{2.32} \\
 & 150K & 18.35 & 14.69 & 0.755 & 2.31 \\
 & 200K & 18.44 & \textbf{14.21} & 0.738 & 2.31 \\
\midrule
\multirow{4}{*}{\texttt{MAR}} 
 & GT   & 54.42 & 55.23 & 0.732 & 1.88 \\
 \cdashline{2-6}
 & \gc100K & \gc\textbf{46.43} & \gc\textbf{41.90} & \gc\textbf{0.607} & \gc\textbf{2.20} \\
 & 150K & 49.75 & 43.42 & 0.527 & 2.15 \\
 & 200K & 52.36 & 45.37 & 0.461 & 2.06 \\
\bottomrule
\end{tabular}
}
\end{table}

Having established the competitiveness of Habibi against a leading commercial system (\S\ref{subsec:11labs_compr_results}), we now turn to a detailed internal analysis. We first examine dialect-specific models to identify per-dialect optimal checkpoints, then compare them against unified models to assess the trade-offs between specialization and generalization. All experiments from this point onward use the full Habibi benchmark with its complete set of zero-shot inference pairs, enabling a thorough assessment across diverse speakers and utterances.

\textbf{Dialect-specific models}\quad Specialized models are trained on D1, with the exception of \texttt{SAU}, which uses the full SADA corpus. As shown in Table~\ref{tab:results_specific}, all specialized models outperform the ground truth (GT) in WER. We attribute this primarily to the fact that GT samples, sourced from ASR corpora, contain higher noise levels than synthesized outputs, which adversely affects ASR-based evaluation. For each dialect, we select the checkpoint with the lowest WER-O for subsequent comparisons.

\textbf{Unified vs.\ specialized models}\quad As shown in Table~\ref{tab:results_unified}, the unified model (Uni.D2-I) achieves performance close to, and in several cases surpassing, that of specialized counterparts. On WER, specialized models retain an edge on \texttt{SAU}, \texttt{UAE}, \texttt{ALG}, \texttt{IRQ}, and \texttt{EGY}, while the unified model is stronger on \texttt{MSA} and \texttt{MAR}. On SIM, the unified model matches or exceeds specialized models on all seven dialects, with the largest gain on \texttt{MAR} (0.705 vs.\ 0.607). UTMOS slightly favors specialized models across most dialects, but remains comparable overall. These results demonstrate that a single unified model can achieve competitive zero-shot synthesis across diverse Arabic dialects, while specialized models remain preferable when maximizing WER on a specific dialect is the priority.

\begin{table*}[ht]
\centering
\caption{Comparing unified models trained on D1 (Uni.D1) and on D2 (Uni.D2) with specialized (Special.) counterparts. The ``-I'' postfix indicates the application of a regional identifier during both training and inference.}
\label{tab:results_unified}
\setlength{\tabcolsep}{3pt}
\resizebox{\textwidth}{!}{
\begin{tabular}{l ccc|ccc|ccc|ccc}
\toprule
\multirow{2}{*}{\textbf{ID~~~~~~~~~~}} 
 & \multicolumn{3}{c}{\textbf{WER-O} $\downarrow$} 
 & \multicolumn{3}{c}{\textbf{WER-S} $\downarrow$} 
 & \multicolumn{3}{c}{\textbf{SIM}   $\uparrow$} 
 & \multicolumn{3}{c}{\textbf{UTMOS} $\uparrow$} \\
\cmidrule(lr){2-4} \cmidrule(lr){5-7} \cmidrule(lr){8-10} \cmidrule(lr){11-13}
  & Special. & Uni.D1-I & Uni.D2-I  & Special. & Uni.D1-I & Uni.D2-I & Special. & Uni.D1-I & Uni.D2-I  & Special. & Uni.D1-I & Uni.D2-I \\ 
\midrule
\texttt{MSA}  
& ~~7.88 & ~~\textbf{7.44} & ~~\underline{7.71}  
& ~~7.74 & ~~\textbf{7.61} & ~~\underline{7.62}  
& ~~0.763 & ~~\textbf{0.767} & ~~\underline{0.764} 
&  ~~~\textbf{1.93} & ~~~\textbf{1.93} & ~~~1.92 \\
\texttt{SAU}  
& \textbf{13.56} & \underline{13.75} & 13.96  
& \textbf{10.42} & 11.16 & \underline{11.13} 
& ~~0.710 & ~~\underline{0.715} & ~~\textbf{0.717}  
& ~~~\textbf{1.71} & ~~~1.66 & ~~~\underline{1.68} \\
\texttt{UAE}  
& ~~\textbf{4.97} & ~~\underline{5.04} & ~~5.15  
& ~~\textbf{4.55} & ~~\underline{4.78} & ~~4.88 
& ~~0.817 & ~~\underline{0.828} & ~~\textbf{0.830}  
& ~~~\textbf{2.88} & ~~~2.71 & ~~~\underline{2.75} \\
\texttt{ALG}  
& \textbf{30.98} & 31.68 & \underline{31.18}  
& \textbf{19.85} & \underline{23.57} & 23.58  
& ~~0.740 & ~~\textbf{0.746} & ~~\textbf{0.746} 
& ~~~\textbf{2.63} & ~~~2.43 & ~~~\underline{2.46} \\
\texttt{IRQ}  
& \underline{17.66} & 17.91 & \textbf{17.12}  
& \textbf{11.45} & \underline{11.82} & 11.90  
& ~~0.763 & ~~\textbf{0.765} & ~~\underline{0.764} 
& ~~~\textbf{2.63} & ~~~2.44 & ~~~\underline{2.48} \\
\texttt{EGY}  
& \textbf{18.08} & 19.04 & \underline{18.58}  
& \textbf{15.42} & 16.89 & \underline{16.44}  
& ~~0.769 & ~~\textbf{0.788} & ~~\underline{0.787}  
& ~~~\textbf{2.32} & ~~~2.12 & ~~~\underline{2.14} \\
\texttt{MAR}  
& 46.43 & \underline{40.12} & \textbf{40.02}  
& \underline{41.90} & 42.63 & \textbf{41.53}  
& ~~0.607 & ~~\textbf{0.705} & ~~\textbf{0.705} 
& ~~~\textbf{2.20} & ~~~1.96 & ~~~\underline{1.98} \\
\bottomrule
\end{tabular}
}
\end{table*}

\subsection{Effectiveness of Curriculum Learning}
\label{subsec:effect_stage_training}

To answer \textbf{RQ1} (\S\ref{sec:experiments}), we compared three training trajectories: (i) training from scratch on dialectal data alone, (ii) direct fine-tuning from the pre-trained Chinese-English base model to dialectal data, and (iii) our proposed two-stage curriculum, which first fine-tunes on MSA before adapting to dialectal distributions.

\begin{table}[ht]
\centering
\caption{Effectiveness of curriculum learning. The settings are defined by the inclusion of a Chinese \& English pretrained model initialization, first-stage MSA-based SFT, and final-stage SFT with a selected corpus. The mark “$\hookrightarrow$ \textit{Cont.}” indicates continued training.}
\label{tab:effect_curriculum}
\setlength{\tabcolsep}{3pt}
\resizebox{\linewidth}{!}{
\begin{tabular}{ccccccc}
\toprule
\textbf{ZH-EN} & \textbf{MSA} & \textbf{Final} & \multirow{2}{*}{\textbf{WER-O}$\downarrow$} & \multirow{2}{*}{\textbf{WER-S}$\downarrow$} & \multirow{2}{*}{\textbf{SIM}$\uparrow$} & \multirow{2}{*}{\textbf{UTMOS}$\uparrow$} \\
\textbf{Init.} & \textbf{SFT} & \textbf{SFT} & & & & \\
\midrule
\multicolumn{3}{c}{Ground Truth} & 28.36 & 19.88~ & 0.724 & 1.71 \\
\hdashline
\xmark & \xmark & SADA &55.63 & 47.54~ & 0.638 & 1.60 \\
\cmark & \xmark & SADA &14.69 & 10.83~ & 0.704 & \textbf{1.72} \\
\cmark & \cmark & SADA & \textbf{13.30} & \textbf{10.20}~ & \textbf{0.709} & 1.71 \\
\midrule
\multicolumn{3}{c}{\quad Ground Truth (avg.)} & 28.27  & 23.01~  & 0.768  & 2.17  \\
\hdashline
\cmark & \xmark & D1 &20.92 & 18.66~ & 0.758 & \textbf{2.27} \\
 & \multicolumn{2}{r}{$\hookrightarrow$ \textit{Cont.}} & 20.39 & 17.34~ & 0.757 & 2.16 \\
\cmark & \cmark & D1 & \textbf{19.50} & \textbf{17.20}~ & \textbf{0.759} & 2.19 \\
\bottomrule
\end{tabular}
}
\end{table}

As shown in the upper half of Table~\ref{tab:effect_curriculum}, training from scratch fails to converge to competitive performance, confirming that cross-lingual transfer from the base model is essential. The curriculum approach consistently outperforms direct fine-tuning across all four metrics, demonstrating that the intermediate MSA stage establishes foundational Arabic phonological and grammatical structure that direct dialectal fine-tuning alone cannot effectively acquire.

We further verified that this conclusion generalizes to the unified modeling setting (lower half of Table~\ref{tab:effect_curriculum}). Here, we additionally compare against a model that receives twice as many dialectal training updates but without the MSA stage (``$\hookrightarrow$ \textit{Cont.}''). Despite having double the gradient updates, this variant still underperforms our curriculum approach, indicating that the \textit{ordering} of training data matters more than the \textit{quantity} of dialectal compute. The consistent pattern across both specialized and unified settings confirms the MSA-first curriculum as a general and effective strategy for multi-dialect Arabic TTS.

\subsection{Effectiveness of In-Context Learning}
\label{subsec:effect_in_context_learning}

To investigate \textbf{RQ2} (§\ref{sec:experiments}), we compared two inference configurations: (i) standard inference with reference speech-text context (\textit{w/} Context in Table~\ref{tab:effect_icl}), and (ii) inference with the reference audio zeroed out, disrupting all prior contextual information (\textit{w/o} Context).

As shown in Table~\ref{tab:effect_icl}, removing the reference context leads to consistent WER degradation across all seven dialects under both WER-O and WER-S. For instance, UAE WER-S increases from 4.88 to 10.68, and IRQ WER-O from 17.12 to 20.15. This confirms that the model actively leverages speech-text cues from the reference to guide dialect-appropriate generation at inference time.

\begin{table}[ht]
\centering
\caption{Effectiveness of in-context learning during the inference of the unified dialectal TTS trained on D2 with regional identifiers (Uni.D2-I), as an example.}
\label{tab:effect_icl}
\setlength{\tabcolsep}{3.5pt}
\resizebox{\linewidth}{!}{
\begin{tabular}{lccccccc}
\toprule
\textbf{Uni.D2-I} & \textbf{\texttt{MSA}} & \textbf{\texttt{SAU}} & \textbf{\texttt{UAE}} & \textbf{\texttt{ALG}} & \textbf{\texttt{IRQ}} & \textbf{\texttt{EGY}} & \textbf{\texttt{MAR}} \\
\midrule
\multicolumn{8}{l}{\textbf{WER-O} $\downarrow$} \\
\textit{w/o} Context  & 13.77 & 18.73 & 12.33 & 32.87 & 20.15 & 18.76 & 42.42 \\
\textit{w/~~} Context & ~~\textbf{7.71} & \textbf{13.96} & ~~\textbf{5.15} & \textbf{31.18} & \textbf{17.12} & \textbf{18.58} & \textbf{40.02} \\
\cdashline{1-8}\noalign{\vskip\belowrulesep}
\multicolumn{8}{l}{\textbf{WER-S} $\downarrow$} \\
\textit{w/o} Context  & 13.30 & 16.37 & 10.68 & 26.87 & 14.63 & 23.22 & 43.09 \\
\textit{w/~~} Context & ~~\textbf{7.62} & \textbf{11.13} & ~~\textbf{4.88} & \textbf{23.58} & \textbf{11.90} & \textbf{16.44} & \textbf{41.53} \\
\bottomrule
\end{tabular}
}
\end{table}

\subsection{Ablation of Data Mixing and Scaling}
\label{subsec:ablation_data}

As described in \S\ref{subsec:dataset}, we applied a denoising enhancement model to the high-noise portion of our training data. 
To assess its impact, we conducted an ablation study on SADA, the dataset used to train the Saudi-specialized TTS model. Specifically, we varied the sampling ratio between original noisy audio and denoised versions during training under three configurations: (i) original audio only (Ratio 0), (ii) denoised audio only (Ratio 1), and (iii) mixed sampling, where the denoised version is selected with probability 0.618.

\begin{table}[ht]
\centering
\caption{Ablation results of data mixing ratios on SADA during Saudi-specialized model training. Evaluated on original (ratio 0) and fully denoised (ratio 1) test sets.}
\label{tab:ablation_data_mixing}
\setlength{\tabcolsep}{3.5pt}
\resizebox{\linewidth}{!}{
\begin{tabular}{l cccccccc}
\toprule
\multirow{1}{*}{\textbf{\texttt{SAU}}} & \multicolumn{2}{c}{\textbf{WER-O} $\downarrow$} & \multicolumn{2}{c}{\textbf{WER-S} $\downarrow$} & \multicolumn{2}{c}{\textbf{SIM} $\uparrow$} & \multicolumn{2}{c}{\textbf{UTMOS} $\uparrow$} \\
\cmidrule(lr){2-3} \cmidrule(lr){4-5} \cmidrule(lr){6-7} \cmidrule(lr){8-9}
\textbf{Ratio} & 0 & 1 & 0 & 1 & 0 & 1 & 0 & 1 \\ 
\midrule
0     & \textbf{13.30} & \textbf{14.33} & \textbf{10.20} & \underline{10.52} & \underline{0.709} & 0.700 & 1.71 & 1.82 \\
0.618 & \underline{13.56} & \textbf{14.33} & \underline{10.42} & \textbf{10.51} & \textbf{0.710} & \textbf{0.702} & 1.71 & \underline{1.90} \\
1     & 14.13 & 14.58 & 10.67 & 10.69 & 0.704 & \textbf{0.702} & \textbf{1.73} & \textbf{1.91} \\
\bottomrule
\end{tabular}
}
\end{table}

\begin{table*}[ht]
\centering
\caption{Comprehensive comparison between unified dialectal models trained with or without regional identifiers (Uni.D1 and Uni.D1-I), and with different inference patterns (plain text, ID-agnostic, and ID-aware) for the latter.}
\label{tab:ablation_regional_id}
\setlength{\tabcolsep}{2.5pt}
\resizebox{\textwidth}{!}{
\begin{tabular}{l cccc|cccc|cccc|cccc}
\toprule
\multirow{3}{*}{\textbf{ID}~~~~~} 
& \multicolumn{4}{c}{\textbf{WER-O} $\downarrow$} 
& \multicolumn{4}{c}{\textbf{WER-S} $\downarrow$} 
& \multicolumn{4}{c}{\textbf{SIM} $\uparrow$} 
& \multicolumn{4}{c}{\textbf{UTMOS} $\uparrow$} \\
\cmidrule(lr){2-5} \cmidrule(lr){6-9} \cmidrule(lr){10-13} \cmidrule(lr){14-17}
 & Uni.D1 & \multicolumn{3}{c}{Uni.D1-I} 
 & Uni.D1 & \multicolumn{3}{c}{Uni.D1-I} 
 & Uni.D1 & \multicolumn{3}{c}{Uni.D1-I} 
 & Uni.D1 & \multicolumn{3}{c}{Uni.D1-I} \\ 
 & $\quad~\cdot\quad$ 
 & $\quad~\cdot\quad$ & $\circledzero\langle~\cdot~\rangle$ & $\circledi\langle~\cdot~\rangle$ 
 & $\quad~\cdot\quad$ 
 & $\quad~\cdot\quad$ & $\circledzero\langle~\cdot~\rangle$ & $\circledi\langle~\cdot~\rangle$ 
 & $\quad~\cdot\quad$ 
 & $\quad~\cdot\quad$ & $\circledzero\langle~\cdot~\rangle$ & $\circledi\langle~\cdot~\rangle$ 
 & $\quad~\cdot\quad$ 
 & $\quad~\cdot\quad$ & $\circledzero\langle~\cdot~\rangle$ & $\circledi\langle~\cdot~\rangle$ \\ 
\midrule
\texttt{MSA} & ~~7.81 & ~~7.64 & ~~\underline{7.51} & \gc~~\textbf{7.44} 
& ~~\underline{7.59} & ~~7.60 & ~~\textbf{7.57} & \gc~~7.61 
& 0.765 & \textbf{0.767} & \textbf{0.767} & \gc\textbf{0.767} 
& \textbf{1.95} & 1.93 & 1.93 & \gc1.93 \\
\texttt{SAU} & 13.88 & \underline{13.87} & 13.97 & \gc\textbf{13.75} 
& \textbf{11.12} & 11.27 & 11.17 & \gc\underline{11.16} 
& \textbf{0.716} & 0.715 & 0.715 & \gc0.715 
& \textbf{1.68} & 1.66 & 1.66 & \gc1.66 \\
\texttt{UAE} & ~~5.30 & ~~5.19 & ~~\underline{5.11} & \gc~~\textbf{5.04} 
& ~~6.17 & ~~\underline{4.94} & ~~4.99 & \gc~~\textbf{4.78}
& 0.827 & \textbf{0.828} & 0.827 & \gc\textbf{0.828} 
& \textbf{2.72} & 2.71 & \textbf{2.72} & \gc2.71 \\
\texttt{ALG} & 32.52 & \textbf{28.86} & \underline{29.16} & \gc31.68 
& 23.75 & \underline{23.69} & 23.84 & \gc\textbf{23.57}
& \textbf{0.746} & 0.744 & 0.744 & \gc\textbf{0.746} 
& 2.44 & \textbf{2.46} & \textbf{2.46} & \gc2.43 \\
\texttt{IRQ} & 18.11 & 17.98 & \textbf{17.71} & \gc\underline{17.91} 
& 11.87 & 12.06 & \textbf{11.82} & \gc\textbf{11.82} 
& \textbf{0.765} & 0.764 & 0.764 & \gc\textbf{0.765} 
& \textbf{2.46} & \underline{2.45} & \underline{2.45} & \gc2.44 \\
\texttt{EGY} & \underline{18.67} & \textbf{18.60} & 18.97 & \gc19.04 
& 17.38 & 17.31 & \underline{17.21} & \gc\textbf{16.89} 
& 0.787 & 0.787 & 0.787 & \gc\textbf{0.788} 
& \textbf{2.13} & 2.12 & 2.12 & \gc2.12 \\
\texttt{MAR} & 40.21 & \textbf{36.13} & \underline{36.65} & \gc40.12 
& 43.30 & 43.79 & \underline{43.26} & \gc\textbf{42.63} 
& 0.704 & \textbf{0.705} & 0.702 & \gc\textbf{0.705} 
& 1.97 & \textbf{1.98} & \textbf{1.98} & \gc1.96 \\
\bottomrule
\end{tabular}
}
\end{table*}

\begin{table}[ht]
\centering
\caption{Impact of data scaling strategies across average sample length, audio quality, and overall data scale on the \texttt{MAR}, \texttt{EGY}, and full datasets.}
\label{tab:ablation_data_scaling}
\setlength{\tabcolsep}{2pt}
\resizebox{\linewidth}{!}{
\begin{tabular}{lr cccc}
\toprule
\multicolumn{2}{l}{\textbf{Data} \qquad\qquad \textbf{Hours}} & \textbf{WER-O}$\downarrow$ & \textbf{WER-S}$\downarrow$ & \textbf{SIM}$\uparrow$ & \textbf{UTMOS}$\uparrow$ \\
\midrule
\multicolumn{6}{c}{\textbf{Moroccan-Specialized Model}} \\
\midrule
GT\texttt{[MAR]} & -    & 54.42  & 55.23  & 0.732  & 1.88  \\
\hdashline
D1\texttt{[MAR]} & 34 & 46.43 & 41.90 & 0.607 & \textbf{2.20} \\
\multicolumn{2}{l}{+ DarijaTTS-clean ~ 43} & 45.70 & 42.15 & 0.640 & 2.12 \\
+ Darija-S2T & 73 & \textbf{44.57} & \textbf{41.63} & \textbf{0.670} & 2.08 \\
\midrule
\multicolumn{6}{c}{\textbf{Egyptian-Specialized Model}} \\
\midrule
GT\texttt{[EGY-MGB2]} & -    & 22.70  & 18.62  & 0.820  & 2.20  \\
\hdashline
D1\texttt{[EGY]} & 37  & \textbf{18.08} & \textbf{15.42} & 0.769 & \textbf{2.32} \\
+ pseudo clean   & 55  & 18.12 & 15.63 & 0.775 & 2.29 \\
+ noisy          & 103 & 18.29 & 16.04 & \textbf{0.778} & 2.29 \\
\midrule
GT\texttt{[EGY-MASC]}  & -    & ~~6.06  & ~~9.97  & 0.773  & 2.07  \\
\hdashline
D1\texttt{[EGY]} & 37  & ~~\textbf{3.58} & ~~\textbf{5.50} & 0.775 & 2.23 \\
+ pseudo clean   & 55  & ~~3.67 & ~~5.57 & \textbf{0.776} & 2.23 \\
+ noisy          & 103 & ~~3.89 & ~~5.96 & \textbf{0.776} & \textbf{2.24} \\
\midrule
\multicolumn{6}{c}{\textbf{Unified Model} (avg. score)} \\
\midrule
GT\texttt{[all]} & -    & 28.27  & 23.01  & 0.768  & 2.17  \\
\hdashline
D1\texttt{[all]} & 1635 & 19.28 & 16.92 & \textbf{0.759} & 2.18 \\
D2\texttt{[all]} & 1857 & \textbf{19.10} & \textbf{16.73} & \textbf{0.759} & \textbf{2.20} \\
\bottomrule
\end{tabular}
}
\end{table}

The results, summarized in Table~\ref{tab:ablation_data_mixing}, show that training and inference on data from the same source distribution typically yield the best performance. However, mixed training with denoised enhancement achieves comparable results on SIM and UTMOS with raw data, while performing better when pure clean speech is used as the reference prompt.
Based on these results, we adopted mixed sampling with a fixed probability of 0.618 as our default training protocol, applied only to sources with a corresponding denoised version; cleaner datasets were not denoised.

To further address \textbf{RQ3} (§\ref{sec:experiments}), we examined three dimensions of data scaling—utterance length diversity, audio quality, and overall dataset size—through targeted ablations summarized in Table~\ref{tab:ablation_data_scaling}.

\textbf{Length diversity}\quad On \texttt{MAR}, augmenting with Darija-S2T (predominantly 20–30s clips) alongside DarijaTTS-clean (predominantly short segments of a few seconds) improves performance, suggesting that a diverse duration distribution benefits TTS training.

\textbf{Audio quality}\quad On \texttt{EGY}, training on clean D1 alone outperforms adding the MASC \textit{noisy} subset or its pseudo-clean filtered version (filtered at a WER-O threshold of 15\%). This holds on both the standard \texttt{EGY-MGB2} test set and a complementary \texttt{EGY-MASC} test set drawn from MASC to reduce distributional bias. The result confirms that data quality remains the dominant factor for TTS.

\textbf{Dataset size}\quad Expanding from D1 to D2 not only broadens dialect coverage (Table~\ref{tab:datasets_dialect}) but also yields performance gains across metrics (average WER-O: 19.28$\to$19.10; WER-S: 16.92$\to$16.73, UTMOS: 2.18$\to$2.20; Table~\ref{tab:ablation_data_scaling}).

\subsection{Ablation of Regional Identifiers}
\label{subsec:ablation_region_id}

To explore \textbf{RQ4} (§\ref{sec:experiments}), we compared unified models trained with and without regional identifiers, and further tested three inference templates for the identifier-aware variant: (i) plain text without any special tokens, $~\cdot~$, (ii) ID-agnostic template $\circledzero\langle~\cdot~\rangle$ that wraps the text with a token marking unknown dialect, and (iii) ID-aware template $\circledi\langle~\cdot~\rangle$ that includes the correct dialect token, matching the training format.

As shown in Table~\ref{tab:ablation_regional_id}, training with regional identifiers generally improves WER while maintaining SIM and UTMOS. We attribute this to the identifiers making dialectal distributions more separable during training, which in turn helps the model disambiguate dialect-specific phonological patterns during generation. Among inference templates, the model is robust across all three configurations, with full training-inference consistency yields the best overall results.

\section{Conclusion}

In this work, we introduce Habibi, the first open-source framework for unified-dialectal Arabic speech synthesis. 
By leveraging linguistically informed curriculum learning and a rigorous data curation pipeline, we demonstrate that high-quality zero-shot synthesis across diverse, low-resource Arabic dialects is achievable, even when training data is primarily repurposed from ASR corpora, and without requiring text diacritization. 
Empirically, our unified model achieves performance comparable to dialect-specific counterparts and even surpasses them on several dialect subsets under our standardized benchmark. 
Moreover, Habibi outperforms the strongest available commercial baseline, ElevenLabs' Eleven v3 (alpha), across major dialect test sets, highlighting the practical competitiveness of an open-source solution in this long-tail setting.

Beyond overall results, our ablations validate the importance of stage-wise curriculum learning (MSA $\rightarrow$ dialectal data) for stable convergence and improved final quality, and demonstrate that dialect-aware regional identifiers can further improve recognition-based metrics with limited impact on speaker similarity and naturalness. 
We also observe that scaling dialect coverage and incorporating additional, carefully filtered data yield measurable gains, though data quality remains a dominant factor. 
All with careful consideration of combining multilingual and specialized monolingual ASR to comprehensively assess dialectal TTS performance.
Finally, we release the model checkpoints, inference code, and the first standardized multi-dialect Arabic zero-shot TTS benchmark, aiming to bridge the resource gap and provide a solid foundation for future research on unified dialect modeling, evaluation, and robust deployment within the Arabic and hopefully broader multilingual speech community.

\section*{Ethical Considerations}
This work is purely a research project. Habibi is developed from open-source model weights, aiming to fill the gap in the field of speech research regarding a unified dialectal modeling of the Arabic language. Existing speaker identification or authentication models should be directly applicable or transferable to our model. Additionally, we emphasize that the terminology setup in this paper does not reflect any official classification for Arabic.


\begin{acks}
The authors would like to express our gratitude to Hanan Aldarmaki, Hongchuan Zeng, Junlin Li, Peng Wang, and Yifan Yang for the valuable discussions.
\end{acks}

\bibliographystyle{ACM-Reference-Format}
\bibliography{custom}


\nocite{open_arabic_asr_leaderboard}

\end{document}